\documentclass{article}
\usepackage{spconf,amsmath,graphicx,hyperref, booktabs, xcolor, float}
\usepackage{graphicx}
\usepackage{subcaption}
\usepackage{cuted}
\usepackage{capt-of}
\usepackage{booktabs}

\title{Quantifying Trait-State Hierarchy in EEG Foundation Model and Signal Processing Embeddings}
\name{Anusha Madapura$^{*}$\thanks{$^{*}$These authors contributed equally to this work.}, Aditya Kommineni$^{*}$, Shrikanth Narayanan}
\address{Signal Analysis and Interpretation Laboratory, University of Southern California, USA}

\begin{document}
\ninept
\maketitle
\begin{abstract}
Electroencephalography (EEG) measures brain activity through recording electrical potentials of the scalp.
EEG recordings capture two distinct sets of characteristics in the signal: (1) \emph{Traits} -- information that remains relatively constant over the course of a recording session, such as subject identity, age, and sex; (2) \emph{States} -- signatures that encode responses to external stimuli, including visual, auditory, and motor information. 
There is limited understanding of how different EEG modeling features (signal processing and foundation models) encode state and trait characteristics.
In this work, we quantify state- and trait-level features captured by EEG embeddings using a hierarchy metric and kNN purity on Brain-Computer Interface dataset trials, and find that a majority of EEG features form a macro-trait hierarchy (embeddings form trait-level clusters with micro-states within).
Our linear probe results indicate that this hierarchy structure leads to suboptimal state distinguishability, negatively impacting BCI performance of EEG foundation models.
To mitigate this, we explore normalization techniques that reduce trait-specific information in embeddings and provide a $\sim$ 10\% improvement in balanced accuracy for state classification.
Additionally, we observe that when fine-tuning models for downstream state decoding, embedding spaces transform from macro-trait to macro-state hierarchies, which in turn lead to better state decoding.
\end{abstract}

\section{Introduction \& Related Work}

Electroencephalography (EEG) is a non-invasive biomedical imaging technique that measures brain activity through scalp-recorded electrical potentials. 
While EEG was initially utilized for neurological diagnosis ~\cite{Smithii2, DEGANS2024101951} and cognitive monitoring \cite{LEHMANN2007342, 2023alzheimersreview}, there is increasing interest to use it in brain-computer interfaces (BCIs) for motor rehabilitation ~\cite{Mane_2020} and speech neuroprostheses ~\cite{willett2023high}. 
EEG modeling approaches have ranged from signal-processing techniques such as frequency band power, entropy, and signal complexity to deep learning approaches that learn task-relevant representations directly from EEG signals. Recently, EEG foundation models pretrained on large-scale datasets using self-supervised learning have demonstrated promising performance for generalization across diverse tasks~\cite{knowledgeguided2024embc, avramidis2025neural,cbramod_2025,csbrain_2025,biot_2023,reve_2025}. However, there has been limited work exploring the latent structure of EEG embeddings.

Neuroscience literature defines \emph{traits} as individual characteristics that are stable across time, while \emph{states} are transient changes in neural activity associated with tasks or cognitive conditions~\cite{state_trait_2017}. 
Latent trait-state theory provides a framework for decomposing observed variability into stable and dynamic components~\cite{steyer1999latent, steyer2015theory}. 
In the context of EEG, trait-related information reflects persistent physiological characteristics, while state-related information captures temporary changes induced by sensory stimuli, motor activity, or cognitive processes. 
We posit that understanding the organization of \textit{traits} and \textit{states} in EEG embeddings is important for the downstream modeling of EEG for diverse applications, including BCI, clinical, and laboratory settings.
For instance, in a motor-imagery setting, we would prefer the EEG model embeddings to primarily capture state information, whereas for neurological-disorder screening, models must selectively capture trait information.

Prior EEG studies examine individual differences, subject identification and cross subject variability in signals ~\cite{DEMURU2020103748}. 
Other studies on EEG representations investigate the relationship between representational geometry and downstream behavior \cite{ding2021grounding}, while Wagh et al. examined latent-space robustness under distribution shifts \cite{wagh2022evaluatinglatentspacerobustness}. 
More recently, Wu et al. \cite{wu2025hierarchical} introduced a geometric hierarchy metric to quantify trait- and state-dominant organization in EEG embeddings, having identified macro-trait structure in dyadic social interaction data. 
However, whether such hierarchical organization generalizes across EEG feature spaces (signal processing and foundation models), generalizes across different BCI tasks, and its relation to downstream decoding performance remain open questions.

In this work, building upon the geometric hierarchy framework introduced by Wu et al. \cite{wu2025hierarchical}, we systematically investigate trait-state organization across traditional signal processing features and pretrained EEG foundation model embedding spaces across diverse BCI tasks. 
We examine the relationship of measured hierarchy to decoding performance and transformation in hierarchies under task-specific fine-tuning. 
Contributions of this work are as stated below:
\begin{itemize}
    \item \textbf{Hierarchy in EEG Features}: Both signal processing and foundation model EEG features form macro-trait clusters on BCI datasets. Additionally, across different models higher macro-trait structure leads to worser state classification.  
    \item \textbf{Subject Mean-normalization improves state classification}: Noticeable performance improvements ($\sim$10\% Balanced Accuracy) can be obtained through normalizing each subject embeddings with their respective mean for state classification tasks indicating the negative impact of macro-trait structure.
    \item \textbf{Fine-tuning transforms embedding hierarchy}: Fine-tuning models leads to a transformation in embedding structure from macro-trait to macro-state, which partially explains sub-optimal linear probe performance of foundation models.
\end{itemize}

\begin{figure}
    \centering
    \includegraphics[width=\linewidth]{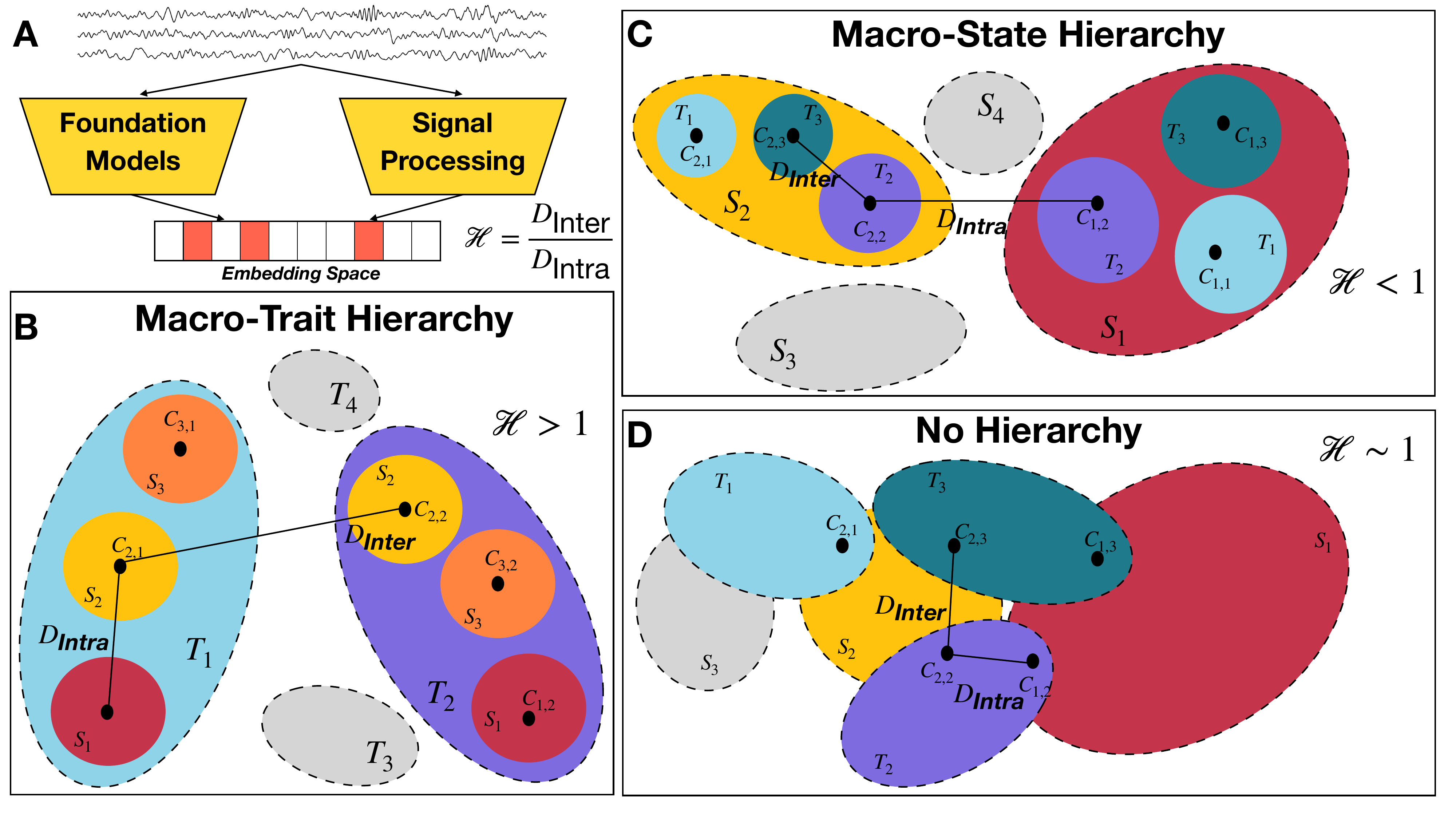}
    \caption{(A) Per-trial embedding computation for foundation models and signal processing features. (B) \textbf{Macro-trait Hierarchy} -- embeddings spaces are grouped based on traits with state specific clusters within, (C) \textbf{Macro-state Hierarchy} -- embeddings clustered based on states with trait specific clusters within and (D) \textbf{No Hierarchy} -- embeddings do not have a macro clustering based on trait or state. $S_{i}$, $T_j$ and $C_{i,j}$ refer to State i, Trait j and Centroid of embeddings for state i and trait j respectively.}
    \label{fig:schematic_hierarchies}
\end{figure}

\section{Methodology}

\subsection{Quantitative Measures of Hierarchy in EEG Features}

Hierarchy of EEG features for trait and state characteristics could take the form of three hierarchical structures as depicted in Fig~\ref{fig:schematic_hierarchies}. (1) \textbf{Macro-Trait Hierarchy} -- embeddings form clusters around traits (subject identities), and within each subject cluster there are micro-state clusters; (2) \textbf{Macro-State Hierarchy} -- embeddings form state clusters with micro-trait clusters within each state cluster; and (3) \textbf{No Hierarchy} -- embeddings do not form any observable macro-state or trait clusters.
We use the \textbf{Hierarchy Metric ($\mathcal{H}$)} and \textbf{kNN Purity} to quantify the hierarchical structure in EEG.

\par The \noindent \textbf{Hierarchy Metric ($\mathcal{H}$)} measures the ratio of inter-trait to intra-trait distances. For each trait $t\in T$ and state $s\in S$, we compute the centroid of the corresponding representations
\begin{equation*}
\mathcal{C}_{s,t} = \frac{1}{N_{s,t}} \sum_{i:\,s_i=s,\,t_i=t} \mathbf{z}_i
\end{equation*}
where $\mathbf{z}_i$ is the representation of observation $i$ and $N{s,t}$ is the number of observations for state $s$ under trait $t$.
Ratio of inter- and intra-trait distances are computed according to the equation below:
\begin{equation*}
    \mathcal{D_{\text{inter}}} = \frac{1}{|S|}\Sigma_{s\in{S}}\frac{1}{|T|.(|T|-1)}\Sigma_{t_i , t_j \in T; t_i \neq t_j} |\mathcal{C}_{s, t_i}-\mathcal{C}_{s, t_j}|_2
\end{equation*}
\begin{equation*}
    \mathcal{D_{\text{intra}}} = \frac{1}{|T|}\Sigma_{t\in{T}}\frac{1}{|S|.(|S|-1)}\Sigma_{s_i , s_j \in S; s_i \neq s_j} |\mathcal{C}_{s_i, t}-\mathcal{C}_{s_j, t}|_2
\end{equation*}
\begin{equation}
    \mathcal{H} = \frac{D_{\text{inter}}}{D_{\text{intra}}}
\end{equation}
\begin{table}[]
    \centering
    \caption{Demographic and recording details for BCI datasets used in evaluation. Length -- length of trials per dataset, SR -- sampling rate and \# Chan -- number of channels in recording}
    \begin{tabular}{lcccc}
        \toprule
        \textbf{Dataset} & \textbf{\# Subjects} & \textbf{SR (Hz)} & \textbf{Length (s)} & \textbf{\# Chan}\\
        \midrule
        Physionet-MI & 105 & 160 & 4 & 64\\
        High Gamma & 14 & 500 & 4 & 128\\
        ERPCore & 40 & 1024 & 1 & 30\\
        \bottomrule
    \end{tabular}
    \label{tab:cohort_sizes}
\end{table}
$D_{\text{intra}}$ measures state-related variation and is calculated as the Euclidean distance between centroids corresponding to different states from the same trait. $D_{\text{inter}}$ measures trait-related variation between individuals and is calculated between centroids corresponding to distinct traits for constant state.  
A ratio $H>1$ indicates a macro-trait hierarchy, in which trait differences dominate state-related variation. Conversely a ratio $H<1$ indicates a macro-state hierarchy, while a $H = 1$ indicates no dominant hierarchical organization.
\par \noindent \textbf{kNN Purity} measures the homogeneity of neighboring embeddings with respect to state and trait labels. 
It is measured at every embedding as the proportion of identical class labels for a given number of neighboring embeddings. 
The value ranges from 0 to 1, with 0 indicating that no neighbors have identical class labels and 1 indicating that all neighbors have identical class labels
For each selected number of neighbors, we report a mean purity value across all embeddings for a given dataset and model.
\begin{table*}[]
    \centering
    \caption{Hierarchy Metric ($\mathcal{H}$) for different BCI datasets across both signal processing and foundation model features. Across all models the hierarchy metric is larger than 1 indicating that model embeddings form macro-trait structure.}
    \begin{tabular}{lccccc|ccc}
    \toprule
     & \multicolumn{5}{c}{\textbf{Foundation Models}} & \multicolumn{3}{c}{\textbf{Signal features}} \\
    \cmidrule(lr){2-6} \cmidrule(lr){7-9}
    \textbf{Dataset} & REVE & BIOT & CSBrain & CBraMod & LaBraM & PSD & Coherence & Complexity \\
    \midrule
    \textbf{Motor-Imagery} \\
    \quad Physionet MI & 4.18 & 5.82 & 5.38 & 5.22 & 5.33 & 3.35 & 4.18 & 6.11 \\
    \quad High Gamma & 5.19 & 3.52 & 3.97 & 3.42 & 4.56 & 4.14 & 4.28 & 6.03 \\
    \midrule
    \textbf{ERPCore } \\
    \quad N400 & 2.57 & 4.61 & 2.73 & 3.12 & 2.81 & 2.55 & 2.58 & 5.88 \\
    \quad flankers & 1.62 & 4.01 & 2.07 & 2.60 & 1.97 & 1.78 & 2.18 & 4.35 \\
    \quad MMN & 5.64 & 18.07 & 7.80 & 11.52 & 7.37 & 7.72 & 6.07 & 20.73 \\
    \quad N170 & 3.27 & 7.90 & 4.12 & 4.54 & 3.43 & 3.39 & 3.46 & 9.69 \\
    \quad N2pc & 5.15 & 12.07 & 7.11 & 8.69 & 5.58 & 4.96 & 4.33 & 13.24 \\
    \quad P3 & 2.47 & 5.49 & 3.40 & 3.64 & 2.74 & 2.69 & 3.04 & 6.70 \\
    \bottomrule
    \end{tabular}
    \label{tab:hierarchy_metric}
\end{table*}
\subsection{EEG Signal Processing and Foundation Model Features}
\label{sec:features}
Hierarchy structures for EEG features are measured both on signal processing and foundation model embeddings. 
Widely used signal processing features include (1) \textbf{Frequency Band Power} -- averaged power spectral density computed from overlapping segments using Welch's method across four EEG frequency bands: theta (4-8 Hz), alpha (8-13 Hz), beta (13-30 Hz), and gamma (30-45 Hz); (2) \textbf{Complexity} -- quantified as the Higuchi Fractal Dimension (HFD) value per channel, which characterizes the irregularity of signal across different temporal scales; and (3) \textbf{Coherence} -- measures the cross correlation between power spectral densities of two EEG channels, with each channel pair producing a scalar value, and is a measure of communication between two lobes of the brain.
While there exists a multitude of EEG foundation models, we pick representative models from different pre-training strategies: \textbf{BIOT}~\cite{biot_2023} for contrastive learning, \textbf{LaBraM}~\cite{labram_2024} for tokenization-based, and \textbf{CBraMod}~\cite{cbramod_2025}, \textbf{CSBrain}~\cite{csbrain_2025} and \textbf{REVE}~\cite{reve_2025} for masked reconstruction models. 
\par \noindent \textbf{Preprocessing pipeline}: Bandpass filter from $0.1 - 50$ Hz, notch filtered (60 Hz) and re-referenced using a common average reference. Bad channels were interpolated during filtering.

\section{Datasets}
\label{sec:datasets}

We analyze BCI datasets, comprising both motor (Physionet-MI, High Gamma) and non-motor (ERPCore dataset composed of six distinct ERP subtasks) paradigms to ensure generalization of findings.
Dataset details are as provided below (Refer to Table~\ref{tab:cohort_sizes}):

\noindent \textbf{Physionet-MI~\cite{schalk2004bci2000}}: Motor movement/imagery dataset comprising EEG recordings from 109 subjects with 64 EEG channels (standard 10-20 montage). 
Each subject completes runs comprising of real/imagined movements of left fist, right fist, both fists or both feet based on the visual cue displayed. 

\par \noindent \textbf{High Gamma Dataset~\cite{schirrmeister2017deep}}: recordings from 14 healthy subjects with a 128-electrode montage. Each subject recording included 1000 four-second trials of real movements. Each trial involves one of four motor movements (left hand, right hand, both feet, or rest). 

\par \noindent \textbf{ERPCore}~\cite{erpcore_2021}: Data collected from 40 healthy young adults covering six distinct ERP paradigms. The entire study consisted of six optimized 10-minute experiments including (1) Face Perception to capture \textbf{N170}, (2) Passive Auditory Oddball -- \textbf{MMN}, (3) Visual Search -- \textbf{N2pc}, (4) Word Pair Judgement -- \textbf{N400}, (5) Active Visual Oddball -- \textbf{P3} and (6) \textbf{Flankers} -- ERN and LRP.
\par 
Across each of the datasets, subject identity is chosen as the trait and task classes are chosen as the states when computing hierarchy metrics and performing downstream classification of states. For instance, in the case of the Physionet-MI dataset when considering data from all subjects, the number of traits would be 109 (equal to the total number of subjects in dataset) and the number of states would be 4 (equal to the number of distinct motor movements).
\section{Experiments}
To test the nature of hierarchical structures formed by EEG features, firstly, we compute the hierarchy metric ($\mathcal{H}$) and kNN Purity values ($k=\{5,10,25,50,100\}$) across the models and datasets stated in Sec~\ref{sec:features} and Sec~\ref{sec:datasets} respectively. 
Then, linear probe experiments for state classification are performed in order to test the correlation between the hierarchy metric and state classification.
Linear probe experiments (frozen foundation model) include a logistic regression model on extracted feature embeddings with a between-subjects train-test ($80$/$20$) split.
Models are trained for $10$ epochs with a learning rate $1e-4$, weight decay of $1e-2$, a batch size $64$ and Adam Optimizer for fine-tuning (foundation model trainable).

\section{Results}
\begin{figure}
    \centering
    \includegraphics[width=\linewidth]{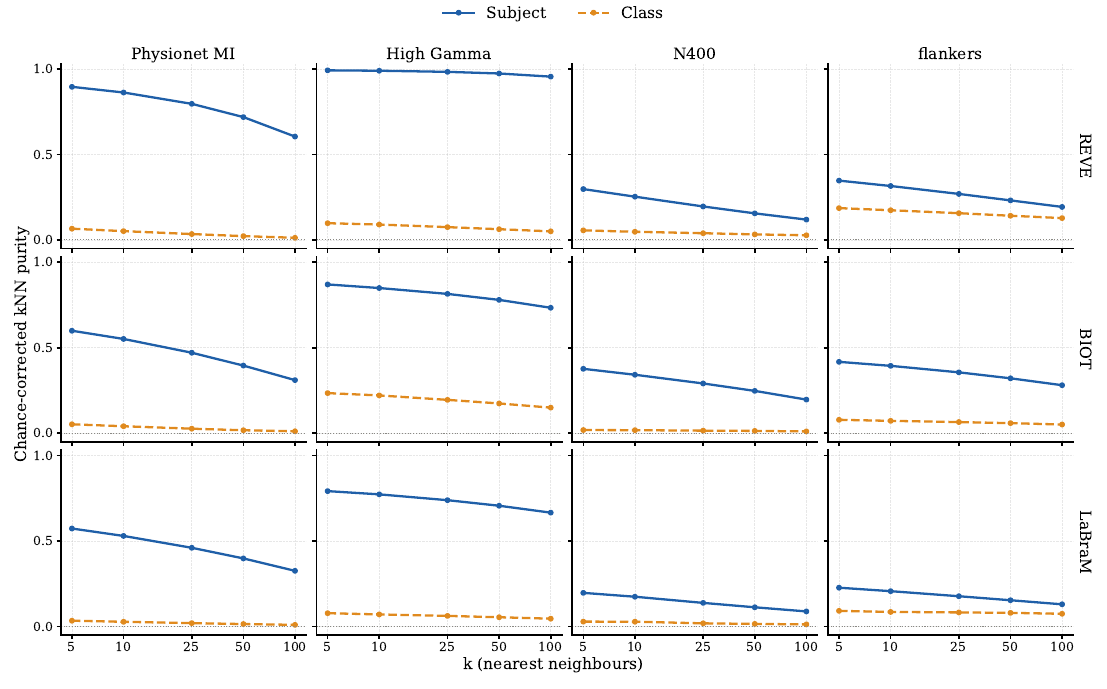}
    \caption{Chance-corrected kNN Purity Plot for REVE (masked reconstruction), BIOT (contrastive) and LaBraM (discrete tokenization). Results indicate higher subject purity compared to state. Plots for some models and datasets were omitted owing to space constraints.}
    \label{fig:knn_purity}
\end{figure}
We study the hierarchical structures of EEG embeddings through probing them both in pre-trained and fine-tuning settings.
In a pre-trained setting, we report hierarchy and kNN purity values across foundation models and signal processing features, and test whether downstream state classification is correlated to hierarchy.
Then, we test whether fine-tuning foundation models results in a transformation in hierarchy from macro-trait to macro-state, which in turn would lead to noticeably better state classification.
\par \noindent \textbf{Do EEG model features have a consistent macro trait/state hierarchy?}
Hierarchy metric values for foundation models and signal processing features across different BCI datasets mentioned in Sec~\ref{sec:datasets} are reported in Table~\ref{tab:hierarchy_metric}. 
Across different models and datasets, we observe a macro-trait organization ($\mathcal{H}{>}1$), with hierarchy values ranging from $\mathcal{H}{=}1.62$ to $\mathcal{H}{=}20.73$. 
Additionally, the contrastive foundation model (BIOT) has noticeably larger hierarchy metrics compared to reconstruction based models, which could be owing to BIOT's pre-training strategy of aligning samples from a given subject together.
These results indicate that across different features, EEG models primarily capture subject identities with micro-clusters within each subject cluster corresponding to state information.

Prior works~\cite{kommineni2026aperiodic} observe that foundation model embeddings provided noticeably better performance for subject decoding as opposed to downstream class decoding.
This phenomenon can be explained from the perspective of hierarchy, as macro-trait structure would lead to better trait decoding since the embeddings form more linearly separable clusters for traits as opposed to states.
While prior work primarily focuses on reconstruction-based foundation models, our results indicate that this phenomenon extends to contrastive EEG foundation models (BIOT) and to EEG signal processing features. 

The kNN purity results in Figure~\ref{fig:knn_purity} further support macro-trait organization. Across neighborhood sizes, trial embeddings exhibit higher subject purity than class purity (when corrected for chance), indicating that nearby embeddings are more likely to originate from the same subject than from the same state. Across datasets and neighborhood sizes, REVE, BIOT, and LaBraM display subject purities above chance of 37.75\%, 39.6\%, and 26.1\%, in contrast to state purities above chance of 2.09\%, 2.43\%, and 1.36\%, respectively. 

We observe that the relative difference between subject and class purity varies across BCI tasks. 
In particular, the Flanker task exhibits a smaller separation between the two purity measures on average, suggesting relative organization of subject and state information depends both on model representations and task. 
One possible explanation is that neural activity during the Flanker task is spatially diffused, which is better captured by pretrained representations. 
\begin{figure}
    \centering
    \includegraphics[width=0.65\linewidth]{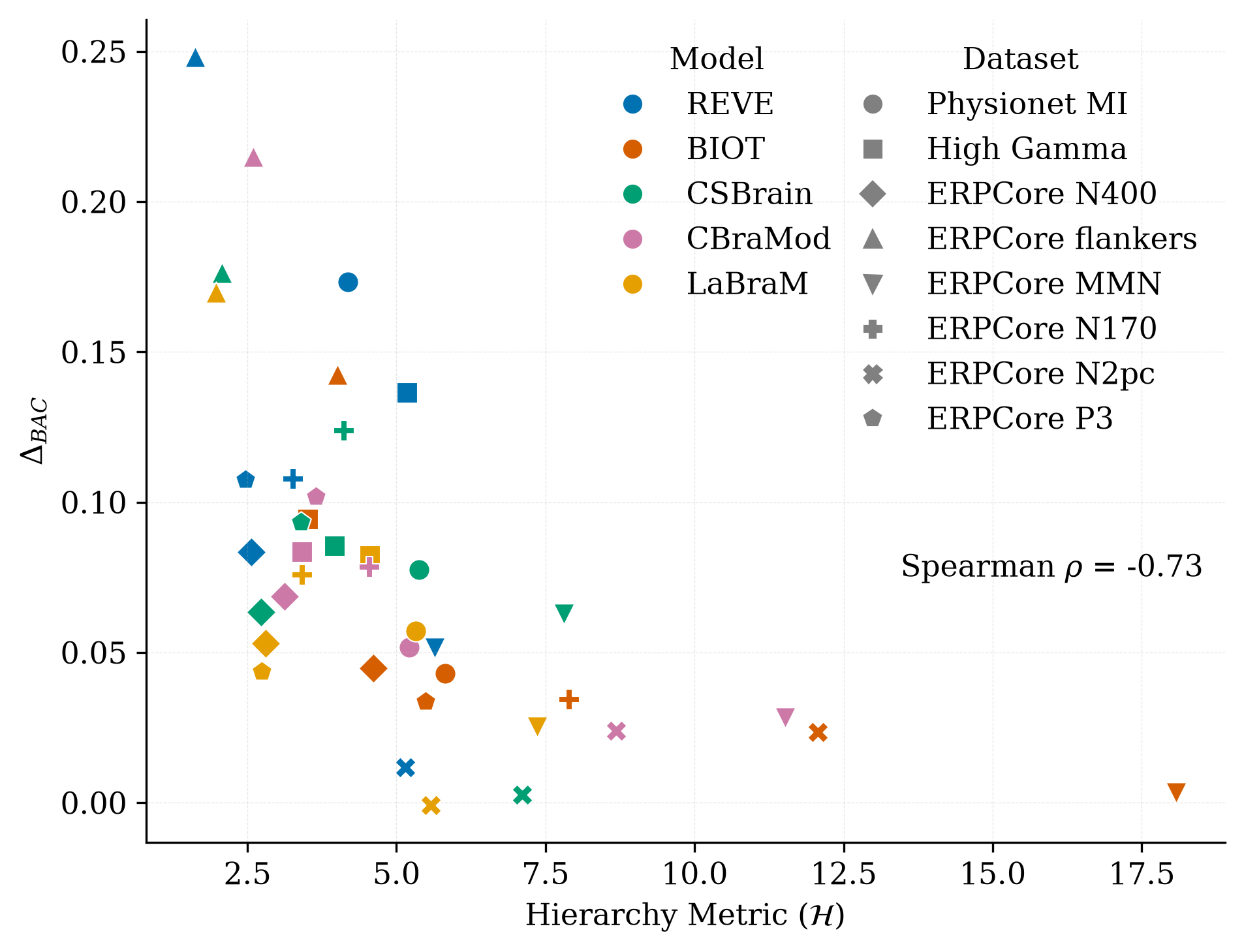}
    \caption{Scatter plot of hierarchy ratio against chance-corrected balanced accuracy ($\Delta_{BAC}$), across models and datasets. Spearmann Correlation of -0.73 indicates a negative correlation between hierarchy ratio and downstream state classification performance.}
    \label{fig:hierarchy_accuracy_linear_probe}
\end{figure}

\noindent \textbf{Does hierarchy metric correlate with downstream state classification?}
To demonstrate the impact of macro-trait structure, we plot the linear-probe state classification performance of foundation models against the hierarchy metric in Figure~\ref{fig:hierarchy_accuracy_linear_probe}.
We observe that higher the hierarchy metric, the lower the state classification performance.
A Spearmann correlation of -0.73 between the hierarchy metric and linear-probe chance-corrected BAC points to macro-trait hierarchy being undesirable for state classification tasks. 
As EEG embeddings form subject-specific macro-clusters, we hypothesize that reducing subject-specific information in embeddings could improve state classification. 
To investigate, we perform subject mean-normalization and evaluate whether state decoding performance improves.

\begin{figure}
    \centering
    \includegraphics[width=0.9\linewidth]{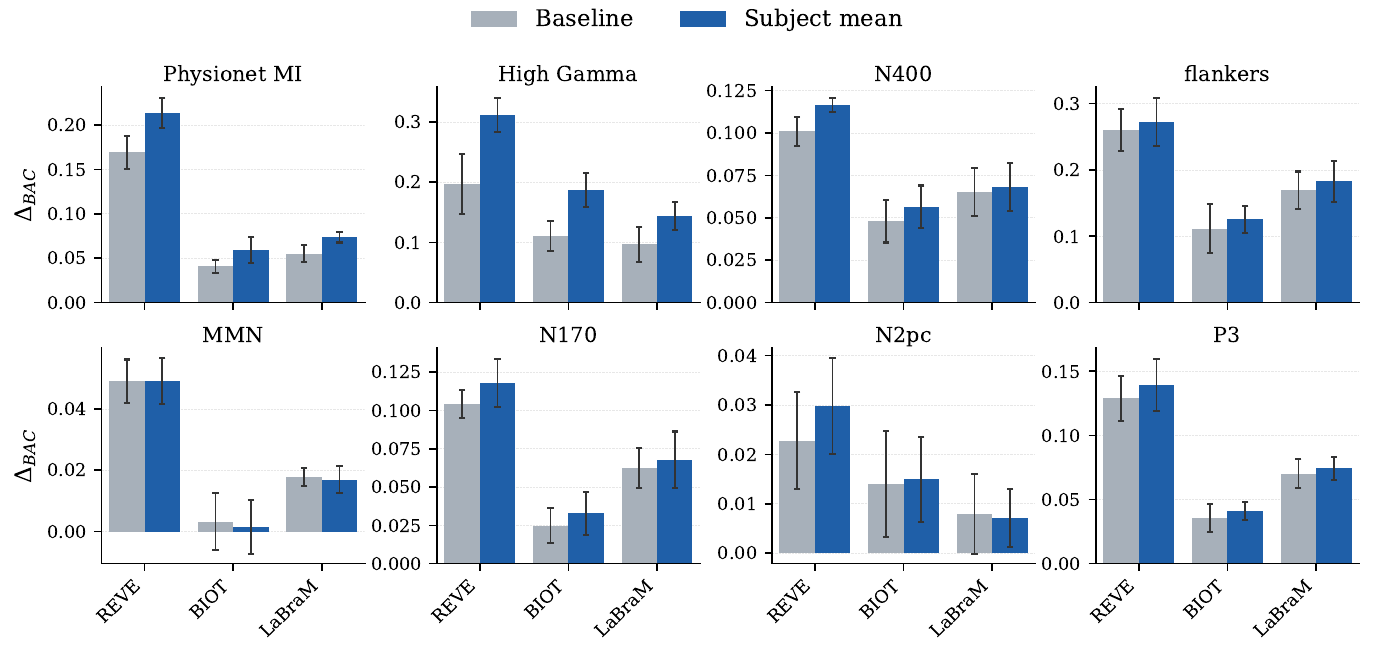}
    \caption{Plot of chance-corrected balanced accuracy ($\Delta_{BAC}$) for baseline embeddings and subject mean-normalized linear probe five-fold classification for state classification reported for REVE, BIOT and LaBraM models. Results indicate subject mean-normalized providing performance gains over baseline which could attribute to reduced subject-identity information in normalized embeddings.}
    \label{fig:subject_mean_results}
\end{figure}

\noindent \textbf{Does modulation of subject-identity information improve state classification performance?}
We compare baseline linear probe performance with subject mean-normalized embeddings, obtained by subtracting each subject's mean embedding. Subject mean normalization would ensure that the signal trial embeddings have reduced subject-specific information.
Five-fold cross-validation results across different datasets and foundation models are reported in Figure~\ref{fig:subject_mean_results}. Compared to baseline performance, subject-normalized models provide performance gains across all datasets, with the largest gains observed in motor-imagery tasks, for PhysionetMI ($5.1$ for REVE) and High Gamma datasets ($11.5$ for REVE). 
ERPCore datasets indicate more modest performance gains, with limited to no gains for tasks that have baseline performance at chance. 

These observations point to two distinct effects: (1) reducing subject-specific information in embeddings provides performance gains, and (2) the magnitude of those gains is dependent on the extent to which the model embeddings capture state-specific information for that task, i.e., the larger state purity within subjects the larger the observed gain with subject normalization. 

\begin{figure}
    \centering
    \includegraphics[width=0.75\linewidth]{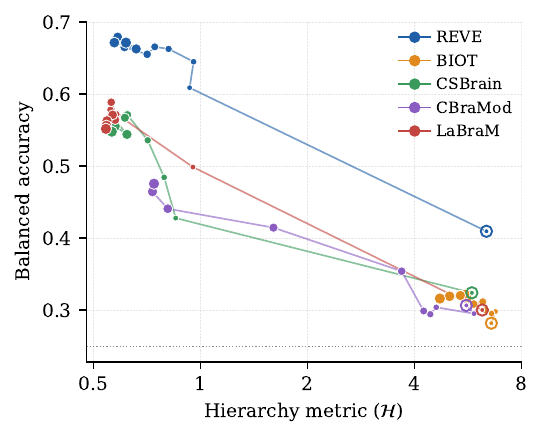}
    \caption{Hierarchy metric ($\mathcal{H}$) plotted against Balanced Accuracy during fine-tuning on physionet-MI dataset. Each data point indicates evaluation at different epochs of fine-tuning}
    \label{fig:finetune_hierarchy_accuracy}
\end{figure}
\noindent \textbf{Does fine-tuning EEG foundation models result in a shift in hierarchy?}
To test whether fine-tuning foundation models shifts the hierarchy structure from macro-trait to macro-state, we fine-tune EEG foundation models (REVE, LaBraM, CSBrain, CBraMod, and BIOT) on the PhysionetMI dataset and study the dynamics of the hierarchy metric during the course of fine-tuning. As Figure~\ref{fig:finetune_hierarchy_accuracy} shows, hierarchy values decrease as fine-tuning progresses, and this is accompanied by an increase in downstream accuracy on state classification. For the best performing models REVE, LaBraM and CSBrain, the hierarchy values drop below 1, indicating a transition from macro-trait ($\mathcal{H}_{\text{REVE}}{=}6.37$, $\mathcal{H}_{\text{LaBraM}}{=}6.20$, $\mathcal{H}_{\text{CSBrain}}{=}5.79$) to macro-state ($\mathcal{H}_{\text{REVE}}{=}0.61$, $\mathcal{H}_{\text{LaBraM}}{=}0.54$, $\mathcal{H}_{\text{CSBrain}}{=}0.56$) structure.

These results suggest that task-specific adaptation alters global subject-state geometry, potentially explaining the suboptimal state-decoding performance of frozen foundation models. Our results demonstrate that EEG representations exhibit predominantly macro-trait organization, which is negatively associated with state decoding performance. 
Subject-mean normalization and task-specific fine-tuning can modify this relationship. 
Therefore, the geometric organization of trait and state information is an important consideration to evaluate EEG representations on downstream tasks.

\section{Discussion \& Conclusion}
Results for the hierarchy metric presented in this work show that both signal processing and EEG foundation model embeddings form a macro-trait hierarchy wherein the embeddings primarily capture subject identity over task states.
While we demonstrate that post-hoc normalization techniques are able to provide performance gains over base embeddings for state classification, our findings, alongside prior interpretability works for EEG foundation models, point to limitations in the pre-training objectives that current EEG foundation models are trained on, thereby leading to embeddings primarily capturing subject identity. 
Future work in EEG foundation modeling should attempt to learn disentangled representations that can capture trait and state representations in distinct subspaces, so as to enable better state classification, especially for BCI tasks.

\bibliographystyle{IEEEbib}
\bibliography{main}
\end{document}